\documentclass[letterpaper]{article} 
\usepackage[preprint]{aaai2027}  
\providecommand{\equalcontrib}{\textsuperscript{*}}
\providecommand{\corresponding}{\textsuperscript{\ensuremath{\dagger}}}
\usepackage[hyphens]{url}  
\usepackage{graphicx} 
\usepackage{natbib}  
\usepackage{caption} 
\usepackage{algorithm}
\usepackage{algorithmic}
\usepackage{amsmath}
\usepackage{multirow}
\usepackage{newfloat}
\usepackage{listings}
\DeclareCaptionStyle{ruled}{labelfont=normalfont,labelsep=colon,strut=off} 
\floatstyle{ruled}
\newfloat{listing}{tb}{lst}{}
\floatname{listing}{Listing}

\usepackage{booktabs}
\usepackage{threeparttable}
\usepackage{colortbl}
\definecolor{OursRow}{RGB}{235,243,250}
\definecolor{ListingBG}{RGB}{238,244,250}

\title{EliSeg: Verified Target Construction for Report-Grounded Abnormality Segmentation}
\author{
    Chengyi Peng\textsuperscript{\rm 1}\equalcontrib,
    Haoyu Yang\textsuperscript{\rm 1}\equalcontrib,
    Meixing Shi\textsuperscript{\rm 1},
    Yuxiang Cai\textsuperscript{\rm 1},
    Yankai Jiang\textsuperscript{\rm 1,\rm 2}\corresponding
}
\affiliations{
    \textsuperscript{\rm 1}Zhejiang University\\
    \textsuperscript{\rm 2}Shanghai Artificial Intelligence Laboratory\\
    chengyipeng0423@home.hpu.edu.cn, yanghaoyu@zju.edu.cn,\\
    shimeixing@zju.edu.cn, caiyuxiang@zju.edu.cn, jyk1996ver@zju.edu.cn
}

\begin{document}

\maketitle

\begin{abstract}

Radiology reports describe clinical observations but do not specify
executable segmentation targets. They may contain present, negated, prior,uncertain, or irrelevant findings, while multiple valid abnormalities may coexist. Existing segmentation methods largely bypass this ambiguity by receiving a target identity or spatial prompt before inference, which acts as a hidden target oracle. We study report-grounded abnormality segmentation, where a model must determine target eligibility, cardinality, and finding-to-mask correspondence directly from an unfiltered report before delineating the corresponding regions. We propose \textbf{EliSeg}, an atcor--verify--revise framework that integrates target construction with mask generation. A grammar-constrained Actor proposes target slots and masks, an independent text-only Verifier reconstructs the eligible finding inventory, and Revision selectively re-executes the shared Actor when their target structures disagree. EliSeg requires no predefined target identity, finding prompt, point, or bounding box. Experiments on MIMIC-CXR-ILS show that EliSeg consistently outperforms direct segmentation methods and extract-then-segment cascades across findings, while effectively suppressing masks for ineligible report mentions. Ablation studies confirm the complementary roles of verification and revision, and evaluation on CheXlocalize demonstrates effective transfer of the EliSeg to an external dataset. Code is available at \url{https://github.com/Maybach-dream/EliSeg}.

\end{abstract}


\begin{figure*}[t]
  \centering
  \includegraphics[width=\textwidth]{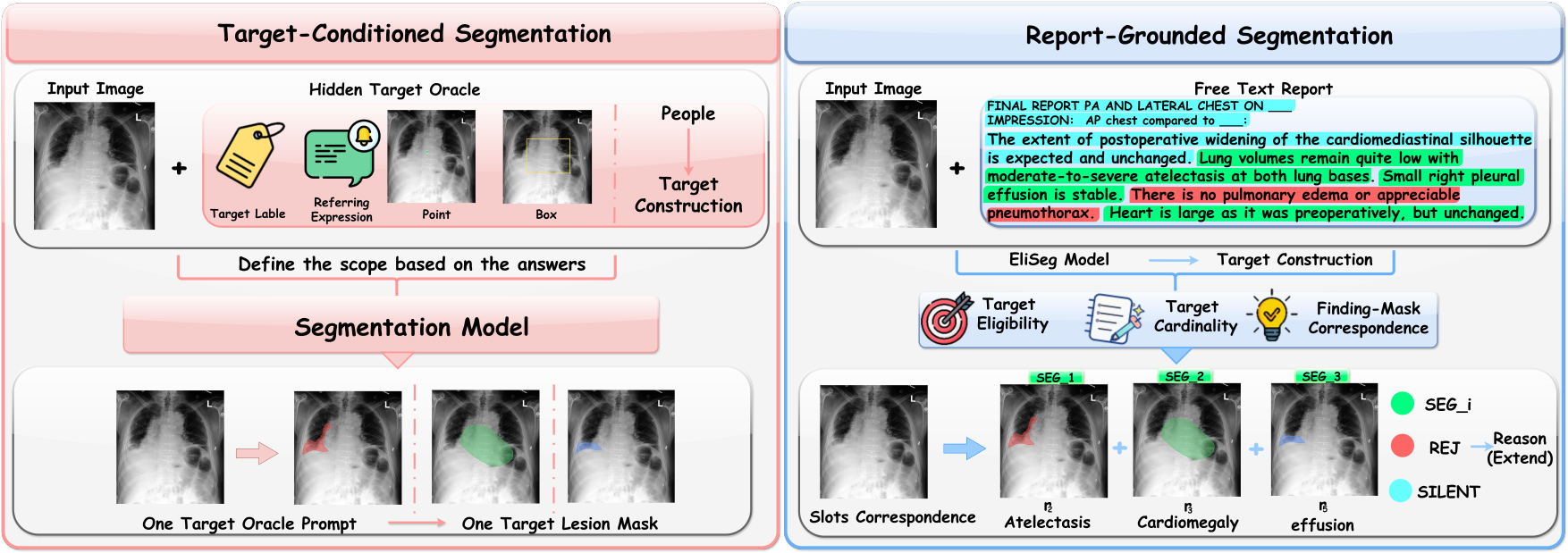}
\caption{Target-conditioned versus report-grounded abnormality segmentation.
The former assumes a hidden target oracle, whereas the latter constructs
eligible targets directly from an unfiltered report.}
  \label{fig:motivation}
\end{figure*}

\section{Introduction}

Chest radiography is widely used for cardiopulmonary assessment, with the
accompanying radiology report recording the radiologist's interpretation of
the image. Establishing pixel-level correspondence between reported
abnormalities and image regions can provide explicit spatial evidence for
clinical findings, supporting image--report consistency checking, lesion
quantification, and clinical quality assurance. Although grounded report
generation methods have begun to associate clinical descriptions with image
regions~\cite{bannur2024maira2}, identifying which findings in an unfiltered
radiology report warrant segmentation and generating a corresponding mask for
each remains insufficiently explored.

Radiology reports are written for clinical communication rather than as
segmentation prompts. They may describe findings that are present, negated,
uncertain, historical, resolved, or outside the segmentation
scope~\cite{irvin2019chexpert,chexbert,jain2021radgraph}. Multiple valid
abnormalities may also coexist and spatially overlap. Consequently, the
occurrence of a disease term does not by itself define a segmentation target.
The report also does not specify how many masks should be generated or which
finding each mask should represent.

We study \emph{report-grounded abnormality segmentation}, where a model
receives a chest radiograph and its associated report evidence, identifies
the findings that warrant spatial localization, and generates a corresponding
mask for each. Unlike target-conditioned segmentation, this setting provides
no preselected finding category, referring expression, target list, point, or
bounding box. The model must therefore construct the eligible target set from
the report before delineating the corresponding image regions.

Most existing segmentation methods assume that the target has already been
specified. Promptable methods receive points or bounding boxes~\cite{sam,sam2,ma2024medsam,zhu2024medicalsam2,imisnet}, while
language-guided and referring segmentation methods receive queries centered
on a requested object or abnormality~\cite{lai2024lisa,huang2025medseg,trinh2025prs,rokuss2025voxtell,
medreasoner, xia2024gsva,choi2025rosalia,tkmamba, liu2023clip,gao2025medical, xing2025diff}. Despite their different interfaces, these
inputs determine what should be segmented before inference. Spatial prompts
may additionally reveal the target location and approximate extent. We refer
to this privileged specification as a \emph{hidden target oracle}. Figure~\ref{fig:motivation} contrasts this conventional setting with
report-grounded segmentation. Rather than receiving a predefined target,
the model must infer target eligibility, cardinality, and finding-to-mask
correspondence from an unfiltered report before generating the corresponding
masks.

Removing this oracle exposes a missing stage between report understanding and
mask prediction: \emph{target construction}. This stage determines whether a
reported finding is eligible for segmentation, how many valid targets are
present, and how the resulting semantic slots correspond to the findings.
Errors therefore extend beyond inaccurate boundaries to include false
targets, omitted findings, incorrect target counts, and semantic mismatches.
These errors cannot be reliably corrected from mask geometry alone. A
plausible mask may originate from an ineligible mention, whereas an omitted
target provides no mask for subsequent refinement.

To address this problem, we propose \textbf{EliSeg}, a structured
propose--verify--revise framework that integrates target construction with
mask generation. A grammar-constrained Actor first predicts a sentence-level
action, target cardinality, and provisional masks. Independently, a
report-only Verifier infers the eligible finding inventory without observing
the image or the Actor outputs. A consistency gate then compares their actions
and cardinalities. When they disagree, Revision constructs a corrected
control sequence and re-executes the shared Actor to generate the final masks.
EliSeg can therefore suppress targets derived from ineligible mentions and
recover omitted mask slots without finding prompts, spatial prompts, or a
separate segmentation pathway. Experiments show that EliSeg improves
report-inferred segmentation across findings, reduces false segmentation of
ineligible mentions, and transfers effectively to an external dataset.

Our main contributions are summarized as follows:
\begin{itemize}
\item We formulate report-grounded abnormality segmentation as target
construction followed by spatial delineation, exposing the hidden target
oracle in conventional target-conditioned protocols.

\item We propose EliSeg, a propose--verify--revise framework that corrects
target eligibility and cardinality without target identities or spatial
prompts at inference.

\item We establish an oracle-free evaluation protocol and show that EliSeg
achieves strong report-inferred segmentation, effective rejection of
ineligible mentions, and successful transfer to CheXlocalize.
\end{itemize}

\section{Related Work}

\subsection{Prompt-Driven Medical Segmentation}

Prompt-driven medical segmentation specifies the target through spatial or
semantic cues. MedSAM, Medical SAM 2, and IMIS-Net use points or bounding
boxes to identify the object of interest
~\cite{ma2024medsam,zhu2024medicalsam2,imisnet}. BiomedParse segments
biomedical objects from textual descriptions and predefined object spaces
~\cite{zhao2025biomedparse}, MedCLIP-SAMv2 converts text--image similarity
into spatial pseudo-prompts~\cite{medclipsamv2}, and MedSAM3 supports
concept-guided medical image segmentation~\cite{jiang2026medicalsam3}. In
chest radiography, ROSALIA generates abnormality masks from instructions
centered on a specified finding~\cite{choi2025rosalia}. Reasoning-based
methods further infer segmentation targets or spatial prompts from implicit
clinical queries before mask generation~\cite{medreasoner,trinh2025prs,tong2025mediround,jiang2026ibisagent,xing2026medvl}. Despite their
different interfaces, these methods constrain the target identity, spatial
location, or candidate semantic space before mask prediction. They therefore
focus on segmenting specified targets rather than determining which findings
in an unfiltered report should become segmentation targets.

\subsection{Report-Grounded Localization}

Report-based methods connect radiological text with image evidence at
different levels. CheXbert extracts structured findings and their assertion
states from radiology reports~\cite{chexbert}, while RadGraph represents
clinical entities and relations in a structured graph
~\cite{jain2021radgraph}. CheXagent supports general chest X-ray understanding
and report-related tasks~\cite{chen2024chexagent}. Beyond report
interpretation, BioViL-T learns localized image--text correspondence from
paired radiographs and reports~\cite{bannur2023biovilt}, while MAIRA-2
associates generated report findings with localized image regions
~\cite{bannur2024maira2}. These methods advance report understanding and
visual grounding, but typically produce structured labels, clinical
descriptions, activation regions, or bounding boxes rather than
finding-specific segmentation masks derived directly from an existing
report. Moreover, they do not explicitly determine which report mentions are
eligible for segmentation, how many masks should be generated, or how those
masks should correspond to individual findings. EliSeg addresses this missing
target-construction stage before final mask decoding.

\begin{figure*}[t]
  \centering
  \includegraphics[width=\textwidth]{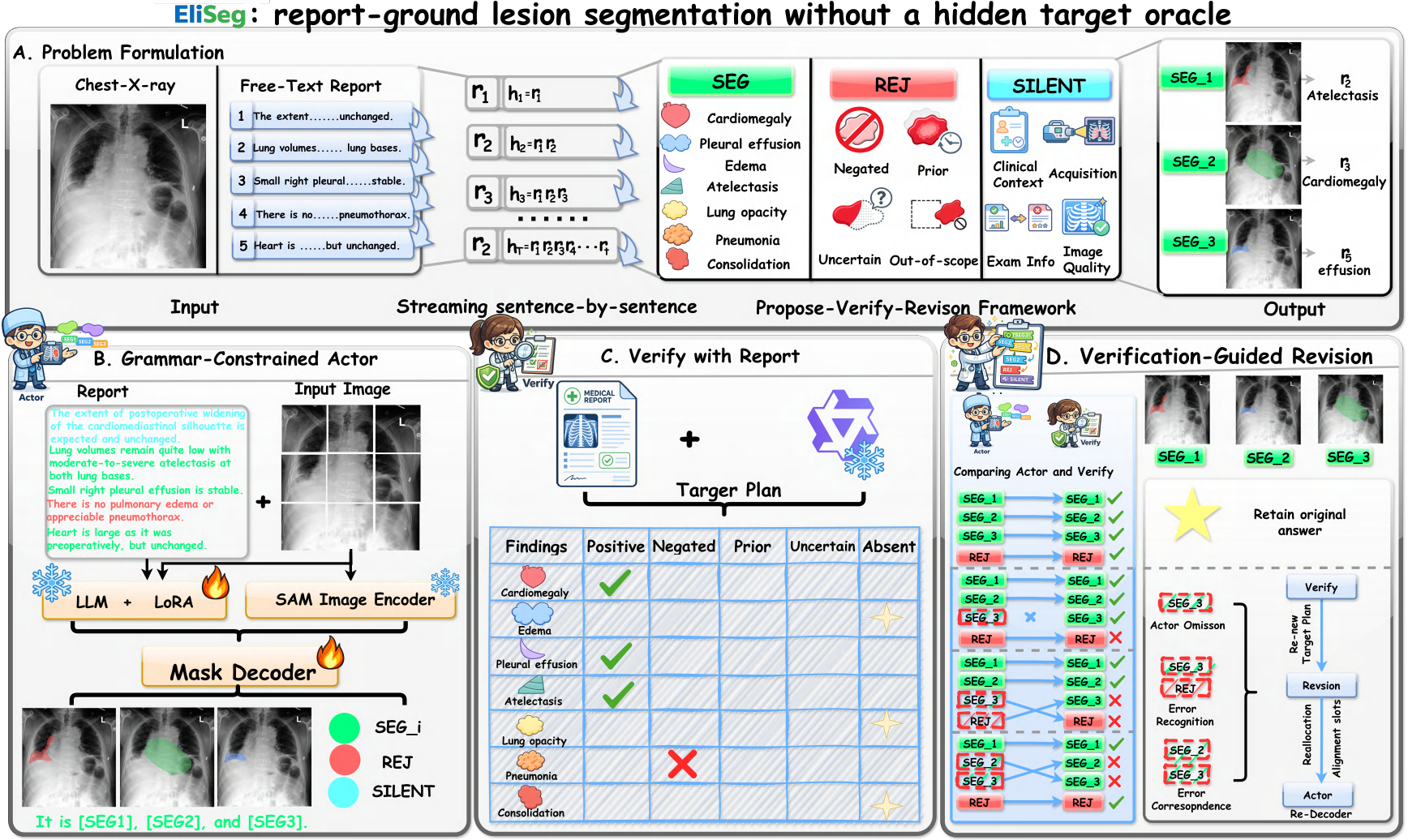}
\caption{Problem formulation and overview of EliSeg.
(A) The model processes each report sentence with its preceding same-section
context and predicts a sentence-level action, together with a set of
finding-specific masks when segmentation is required.
(B) The grammar-constrained Actor proposes an action, target cardinality, and
provisional masks from the image--report context.
(C) The Verifier independently infers the eligible finding
inventory and its cardinality from the report.
(D) A consistency gate retains the Actor output when the Actor and Verifier
agree on action and cardinality; otherwise, Revision constructs a corrected
control sequence and re-executes the shared Actor to generate the final
masks.}
  \label{fig:framework}
\end{figure*}

\section{Method}
\label{sec:method}

\subsection{Problem Formulation}
\label{sec:problem_formulation}

We formulate report-grounded abnormality segmentation as sentence-level target construction followed by spatial delineation. Let \(I\) denote a chest radiograph and \(\mathcal{R}=(r_1,\ldots,r_T)\) its associated radiology report. For each sentence \(r_t\), the model predicts a set of segmentation targets conditioned on \(I\), \(r_t\), and the preceding sentences \(h_t\) from the same report section.

We focus on seven chest radiographic abnormalities: cardiomegaly, edema, pleural effusion, atelectasis, lung opacity, pneumonia, and consolidation. A finding is eligible for segmentation when \((h_t,r_t)\) asserts its current presence. Negated, prior, uncertain, and out-of-scope findings are considered ineligible, where \emph{prior} subsumes both historical and resolved findings.

For sentence \(r_t\), the target set is defined as
\begin{equation}
\mathcal{T}_t
=
\left[
(f_{t,1},M_{t,1}),
\ldots,
(f_{t,K_t},M_{t,K_t})
\right],
\label{eq:target_set}
\end{equation}
where \(K_t\) denotes the target cardinality, \(f_{t,k}\) the semantic identity of the \(k\)-th target, and \(M_{t,k}\) its corresponding binary mask. The semantic category, rather than an anatomical instance or connected component, constitutes the prediction unit. Each category contributes at most one target per sentence; bilateral, multifocal, and diffuse manifestations are merged into a single category-level mask. Distinct categories retain separate masks and may spatially overlap. Target pairs are serialized according to a fixed order over the finding vocabulary.

The sentence-level action
\(a_t\in\{\mathrm{SEG},\mathrm{REJ},\mathrm{SILENT}\}\)
summarizes the target status of \(r_t\). Specifically, \(a_t=\mathrm{SEG}\) if \(K_t>0\), \(a_t=\mathrm{REJ}\) if no eligible target exists but explicit ineligible evidence is present, and \(a_t=\mathrm{SILENT}\) otherwise. When eligible and ineligible findings co-occur, \(\mathrm{SEG}\) takes precedence.

At inference, no preselected target identity, target-specific expression,
target list, point, bounding box, or reference mask is provided. The model
must therefore infer target eligibility, target cardinality, and
finding-to-mask correspondence from the image--report context before
delineating the corresponding regions.

\subsection{EliSeg Overview}
\label{sec:overview}

Figure~\ref{fig:framework} presents the overall architecture of EliSeg.
Given \((I,h_t,r_t)\), the grammar-constrained Actor predicts a sentence-level
action, target cardinality, and provisional masks
(Fig.~\ref{fig:framework}B). Independently, the report-only Verifier infers
the eligible finding inventory from \((h_t,r_t)\), yielding a text-based
action and target cardinality (Fig.~\ref{fig:framework}C).

A deterministic consistency gate compares the actions and cardinalities
predicted by the Actor and Verifier. If they agree, the original Actor output
is retained. Otherwise, Revision constructs a corrected control sequence and
re-executes the shared Actor to generate the final masks
(Fig.~\ref{fig:framework}D). This selective correction preserves consistent
image-grounded predictions without introducing finding prompts, spatial
prompts, or an additional segmentation pathway.

\subsection{Grammar-Constrained Actor}
\label{sec:actor}

The Actor comprises an image encoder \(\mathcal{E}\), a causal language
model, a projection module \(g\), and a mask decoder \(\mathcal{D}\). It realizes the structured prediction defined in Sec.~\ref{sec:problem_formulation} through a compact vocabulary of control tokens. An empty response represents \(\mathrm{SILENT}\), while \([\mathrm{REJ}]\) represents an explicit rejection. For a segmentation response, the Actor emits one control token for each proposed target, yielding an ordered sequence from \([\mathrm{SEG1}]\) to \([\mathrm{SEG}K_t^A]\), where \(K_t^A\) denotes the target cardinality proposed by the Actor. Grammar-constrained decoding restricts the response to a valid structure and
limits the number of segmentation slots to \(K_{\max}=3\). We select this
limit based on the training-set cardinality distribution, where the vast
majority of eligible sentences contain no more than three targets.

Each segmentation token serves as a semantic query for mask decoding. Let
\(z_{t,k}^{A}\) denote the contextual representation associated with the
\(k\)-th segmentation token. The corresponding tentative mask is decoded as
\begin{equation}
\widehat{M}_{t,k}^{A}
=
\mathcal{D}\bigl(\mathcal{E}(I),g(z_{t,k}^{A})\bigr),
\qquad
k=1,\ldots,K_t^A.
\label{eq:actor_mask}
\end{equation}
The Actor does not emit explicit finding labels. Instead, the semantic
identity of each slot is encoded by its report-conditioned representation
and the canonical target ordering used during supervision.

The Actor is jointly optimized for structured response generation and mask
prediction:
\begin{equation}
\mathcal{L}_{\mathrm{Actor}}
=
\mathcal{L}_{\mathrm{CE}}
+
\lambda_{\mathrm{BCE}}\mathcal{L}_{\mathrm{BCE}}
+
\lambda_{\mathrm{Dice}}\mathcal{L}_{\mathrm{Dice}}.
\label{eq:actor_loss}
\end{equation}
\(\mathcal{L}_{\mathrm{CE}}\) supervises the grammar-constrained control
sequence, including the sentence-level action and proposed target cardinality,
whereas \(\mathcal{L}_{\mathrm{BCE}}\) and
\(\mathcal{L}_{\mathrm{Dice}}\) provide pixel-wise and region-level supervision
for masks associated with valid segmentation slots. Reference masks are
assigned to slots according to the canonical vocabulary order, while padded
slots receive zero mask-loss weight.

\subsection{Verifier}
\label{sec:verifier}

Since the Actor may include ineligible findings or omit eligible ones, we
introduce a frozen Qwen2.5-VL-7B model~\cite{bai2025qwen25vl} as an independent Verifier. Given only
\((h_t,r_t)\), it deterministically generates a structured record
under a fixed few-shot prompt, without access to the radiograph or any Actor
outputs.

The record characterizes each target-relevant finding using one of five
textual states: \(\mathrm{positive}\), \(\mathrm{negated}\),
\(\mathrm{prior}\), \(\mathrm{uncertain}\), or \(\mathrm{absent}\).
The \(\mathrm{prior}\) state covers findings described as historical or
resolved rather than currently present, whereas \(\mathrm{absent}\) denotes
that a finding is unmentioned in the current sentence rather than explicitly
negated. Positively asserted findings beyond the segmentation scope are
recorded separately.

Findings assigned \(\mathrm{positive}\) are arranged in canonical vocabulary
order to form the verified inventory \(\mathcal{L}_t^V\), with
\(K_t^V=|\mathcal{L}_t^V|\). This inventory determines the target cardinality
used by Revision. The Verifier operates exclusively on textual evidence and
does not assess reference-mask availability or spatial properties.

\begin{table*}[t]
  \centering
  \begin{threeparttable}
    \footnotesize
    \setlength{\tabcolsep}{3.5pt}
    \renewcommand{\arraystretch}{1}
    \begin{tabular}{lrrrrrrrrrrrr}
      \toprule
      \multicolumn{1}{c}{\multirow{2}{*}{Method}}
      & \multicolumn{4}{c}{
          \shortstack{Report-Inferred\\$\mathrm{R}$}
        }
      & \multicolumn{4}{c}{
          \shortstack{Report + Gold Inventory\\$\mathrm{R+G}$}
        }
      & \multicolumn{4}{c}{
          \shortstack{Native Prompt\\
    $\mathrm{G}_{\mathrm{native}}$}
        } \\
      \cmidrule(lr){2-5}
      \cmidrule(lr){6-9}
      \cmidrule(l){10-13}
      & IoU $\uparrow$
      & Dice $\uparrow$
      & NSD $\uparrow$
      & HD95 $\downarrow$
      & IoU $\uparrow$
      & Dice $\uparrow$
      & NSD $\uparrow$
      & HD95 $\downarrow$
      & IoU $\uparrow$
      & Dice $\uparrow$
      & NSD $\uparrow$
      & HD95 $\downarrow$ \\
      \midrule

      \multicolumn{13}{l}{
        \emph{(a) Direct text-based segmentation}
      } \\

      GSVA
        & 4.6 & 8.8 & 0.3 & 1329.1
        & 3.7 & 7.2 & 0.2 & 1372.1
        & 1.4 & 2.8 & 0.0 & 1422.5 \\

      MedCLIP-SAMv2
        & 13.8 & 24.3 & 7.3 & 370.8
        & 13.0 & 23.1 & 7.0 & 375.4
        & 12.2 & 21.7 & 6.5 & 385.4 \\

      BiomedParse
        & 16.2 & 27.9 & 10.8 & 412.8
        & 16.1 & 27.8 & 10.7 & 426.5
        & 15.3 & 26.6 & 11.1 & 424.7 \\

      MedSAM3
        & 18.0 & 30.6 & 10.7 & 349.9
        & 18.4 & 31.1 & 10.9 & 347.5
        & 18.9 & 31.7 & 11.2 & 355.3 \\

      CheXagent
        & 25.1 & 40.1 & 9.4 & 288.6
        & 25.3 & 40.4 & 9.4 & 298.6
        & 28.0 & 43.8 & 10.0 & 266.5 \\

      MAIRA-2
        & 29.6 & 45.6 & 12.0 & 287.6
        & 26.9 & 42.4 & 9.6 & 541.4
        & 31.1 & 47.4 & 9.9 & 405.8 \\

      ROSALIA
        & 40.8 & 57.9 & 18.1 & 351.2
        & 52.5 & 68.9 & 21.3 & 286.4
        & 56.3 & 72.0 & 23.4 & 229.6 \\

      \addlinespace[2pt]
        \multicolumn{13}{l}{
          \emph{(b) Extract-then-segment cascades}
        } \\
        
        CheXbert$\rightarrow$MedCLIP-SAMv2
          & 11.8 & 21.2 & 6.0 & 458.6
          & \multicolumn{4}{c}{--}
          & \multicolumn{4}{c}{--} \\
        
        CheXbert$\rightarrow$CheXagent
          & 26.9 & 42.4 & 9.3 & 347.4
          & \multicolumn{4}{c}{--}
          & \multicolumn{4}{c}{--} \\
        
        CheXbert$\rightarrow$MAIRA-2
          & 30.3 & 46.5 & 9.3 & 469.3
          & \multicolumn{4}{c}{--}
          & \multicolumn{4}{c}{--} \\
        
        CheXbert$\rightarrow$ROSALIA
          & 53.9 & 70.0 & 21.9 & 311.6
          & \multicolumn{4}{c}{--}
          & \multicolumn{4}{c}{--} \\

      \addlinespace[2pt]
        \multicolumn{13}{l}{
          \emph{(c) Gold-derived spatial prompting}
        } \\
        
        MedSAM (tight box)
          & \multicolumn{4}{c}{--}
          & \multicolumn{4}{c}{--}
          & 47.3 & 64.2 & 22.8 & \textbf{82.0} \\
        
        IMIS-Net (point)
          & \multicolumn{4}{c}{--}
          & \multicolumn{4}{c}{--}
          & 25.6 & 40.7 & 12.0 & 217.1 \\
        
        IMIS-Net (tight box)
          & \multicolumn{4}{c}{--}
          & \multicolumn{4}{c}{--}
          & 48.4 & 65.2 & 22.8 & 82.5 \\

      \addlinespace[2pt]
      \multicolumn{13}{l}{
        \emph{(d) Integrated target construction and segmentation}
      } \\

      \textbf{EliSeg (Ours)}
        & \textbf{59.4} & \textbf{74.5}
        & \textbf{23.2} & \textbf{213.9}
        & \textbf{60.8} & \textbf{75.6}
        & \textbf{24.1} & \textbf{170.4}
        & \textbf{59.4} & \textbf{74.5}
        & \textbf{23.2} & 213.9\\

      \bottomrule
    \end{tabular}

    \caption{Segmentation performance under three input settings. Under
    \(\mathrm{R}\), targets are inferred from the unfiltered report; under
    \(\mathrm{R+G}\), the gold eligible-finding inventory is additionally
    provided without spatial cues; under
    \(\mathrm{G}_{\mathrm{native}}\), each model receives the target through its
    native text or spatial interface. Extract-then-segment cascades are evaluated
    only under \(\mathrm{R}\), whereas gold-derived spatial prompts are evaluated
    only under \(\mathrm{G}_{\mathrm{native}}\). For EliSeg, the native input is
    the unfiltered report, so its \(\mathrm{G}_{\mathrm{native}}\) results
    coincide with those under \(\mathrm{R}\). Dashes denote inapplicable
    settings.}

    \label{tab:main}
  \end{threeparttable}
\end{table*}

\subsection{Verification-Guided Revision}
\label{sec:revision}

Revision converts the verified finding inventory into an executable target
plan. Its target cardinality is revised as
\begin{equation}
\widetilde{K}_t
=
\min\bigl(K_t^V,K_{\max}\bigr),
\label{eq:revised_cardinality}
\end{equation}
A nonempty verified inventory yields a \(\mathrm{SEG}\) action. When no eligible finding is verified, the action is set to \(\mathrm{REJ}\) if the report contains explicit negated, prior, uncertain, or positive out-of-scope evidence, and to \(\mathrm{SILENT}\) otherwise.

For a revised \(\mathrm{SEG}\) action, Revision constructs a canonical control
sequence containing \(\widetilde{K}_t\) segmentation tokens. Rather than
invoking autoregressive decoding again, the complete sequence is supplied to
the shared Actor through an inference-mode, teacher-forced forward pass. The
contextual representation preceding each segmentation token is then decoded
into a final mask using the same image encoder, projection module, and mask
decoder defined in Eq.~\ref{eq:actor_mask}. This re-execution introduces no
gradient update or additional learned parameters.

The verified finding names are not injected into the Actor. The Verifier
inventory determines only the revised action and target cardinality, while
the semantic correspondence of the slots is reconstructed from the
report-conditioned Actor representations under the canonical ordering.
Revision therefore corrects the executable target structure without relying
on target-specific expressions, spatial prompts, or a separate mask-generation
pathway.


\begin{table*}[t]
  \centering
  \begin{threeparttable}
    \small 
    \begin{tabular}{lrrrrrrrrrrrr}
      \toprule
      & \multicolumn{4}{c}{ROSALIA}
      & \multicolumn{4}{c}{CheXbert$\rightarrow$ROSALIA}
      & \multicolumn{4}{c}{EliSeg} \\
      \cmidrule(lr){2-5}
      \cmidrule(lr){6-9}
      \cmidrule(l){10-13}
      Finding
      & IoU $\uparrow$ & Dice $\uparrow$ & NSD $\uparrow$ & HD95 $\downarrow$
      & IoU $\uparrow$ & Dice $\uparrow$ & NSD $\uparrow$ & HD95 $\downarrow$
      & IoU $\uparrow$ & Dice $\uparrow$ & NSD $\uparrow$ & HD95 $\downarrow$ \\
      \midrule

      Cardiomegaly
      & 42.0 & 59.2 & 19.4 & 395.6
      & 81.7 & 89.9 & \textbf{33.8} & 81.3
      & \textbf{84.5} & \textbf{91.6} & 33.0 & \textbf{57.3} \\

      Edema
      & 45.5 & 62.5 & 21.3 & 317.6
      & 52.4 & 68.7 & 23.8 & 330.6
      & \textbf{58.4} & \textbf{73.7} & \textbf{25.0} & \textbf{239.3} \\

      Pleural effusion
      & 33.3 & 49.9 & 16.1 & 394.0
      & 39.6 & 56.7 & 20.0 & \textbf{243.0}
      & \textbf{43.7} & \textbf{60.8} & \textbf{22.9} & 266.8 \\

      Atelectasis
      & 33.4 & 50.0 & 17.8 & 330.3
      & 43.1 & 60.2 & 21.4 & 279.5
      & \textbf{48.2} & \textbf{65.1} & \textbf{22.1} & \textbf{255.0} \\

      Lung opacity
      & 36.1 & 53.0 & 17.2 & 363.4
      & 43.6 & 60.8 & \textbf{19.4} & 372.4
      & \textbf{49.5} & \textbf{66.2} & \textbf{19.4} & \textbf{214.7} \\

      Pneumonia
      & 38.0 & 55.0 & 15.9 & 335.1
      & 36.2 & 53.2 & 16.1 & 446.6
      & \textbf{45.4} & \textbf{62.4} & \textbf{19.5} & \textbf{222.0} \\

      Consolidation
      & 46.0 & 63.0 & 18.6 & 314.7
      & 44.8 & 61.9 & 17.7 & 469.3
      & \textbf{51.5} & \textbf{68.0} & \textbf{19.9} & \textbf{249.0} \\

      \midrule
      \textbf{All}
      & 40.8 & 57.9 & 18.1 & 351.2
      & 53.9 & 70.0 & 21.9 & 311.6
      & \textbf{59.4} & \textbf{74.5} & \textbf{23.2} & \textbf{213.9} \\

      \bottomrule
    \end{tabular}

    \caption{Per-finding segmentation performance under Report-Inferred
    inference. The \emph{All} row aggregates performance over all 1,008
evaluated targets.}

    \label{tab:per_finding}
  \end{threeparttable}
\end{table*}

\section{Experiments}
\label{sec:experiments}

\subsection{Experimental Setup}
\label{sec:experimental_setup}

\paragraph{Evaluation sets.}

We evaluate on MIMIC-CXR-ILS~\cite{johnson2019mimic}, which pairs chest radiographs and radiology
reports with finding-specific segmentation masks. The test set
contains 1,008 finding-level segmentation targets with valid reference
masks, where each target corresponds to one report-supported finding in an
image. We separately evaluate rejection on 600 ineligible report mentions,
balanced across negated, prior, and uncertain evidence. For external
evaluation, we use CheXlocalize~\cite{saporta2022benchmarking}, which provides
expert-annotated abnormality masks but no paired radiology reports, to assess
zero-shot segmentation transfer in a given-target setting. 

\paragraph{Input settings.}

We evaluate three input settings that differ in how segmentation targets are
provided. Under \(\mathrm{R}\), the model receives the radiograph and the
unfiltered report and must construct the targets before segmentation. Under
\(\mathrm{R+G}\), the gold eligible-finding inventory is provided together
with the report, without spatial cues. Under
\(\mathrm{G}_{\mathrm{native}}\), each gold target is provided through the
model's native interface, such as a finding prompt, point, or bounding box.
This setting evaluates each model's segmentation capability through its
native target interface.

\paragraph{Baselines.}

We compare EliSeg with text-based segmentation methods, including GSVA,
MedCLIP-SAMv2, BiomedParse, MedSAM3, CheXagent, MAIRA-2, and
ROSALIA~\cite{xia2024gsva,medclipsamv2,zhao2025biomedparse,
jiang2026medicalsam3,chen2024chexagent,bannur2024maira2,
choi2025rosalia}. We additionally construct
extract-then-segment cascades in which CheXbert~\cite{chexbert} predicts
eligible findings before segmentation; these cascades are evaluated only
under \(\mathrm{R}\). MedSAM and IMIS-Net~\cite{ma2024medsam,imisnet}
serve as spatially prompted baselines and are evaluated using tight
spatial prompting under
\(\mathrm{G}_{\mathrm{native}}\).

\paragraph{Implementation details.}

The Actor is initialized from ROSALIA-7B. Its mask decoder and text
projection module are trainable, while the language backbone is adapted
using rank-8 LoRA~\cite{hu2022lora}. We use
AdamW~\cite{loshchilov2017decoupled} with a learning rate of
\(5\times10^{-5}\), a batch size of 2, and gradient accumulation over 4
steps. Images and masks are resized to \(1024\times1024\). All experiments
are conducted on an NVIDIA RTX A6000 GPU.

\paragraph{Metrics.}

We report intersection over union (IoU), Dice coefficient, normalized surface
Dice (NSD)~\cite{nsd}, and the 95th-percentile Hausdorff distance (HD95). IoU and Dice
are pooled across eligible targets, whereas NSD and HD95 are averaged over
individual targets. Missing, rejected, or failed predictions for eligible
targets are retained and treated as empty masks. For ineligible mentions, we
report the false-segmentation rate (FSR), defined as the proportion producing
a nonempty mask, together with \(\mathrm{IoU}_{+}\), the IoU evaluated on
eligible targets, to measure the trade-off between rejection and segmentation
performance.

\subsection{Main Results}
\label{sec:main_results}

\begin{figure*}[t]
  \centering
  \includegraphics[width=\textwidth]{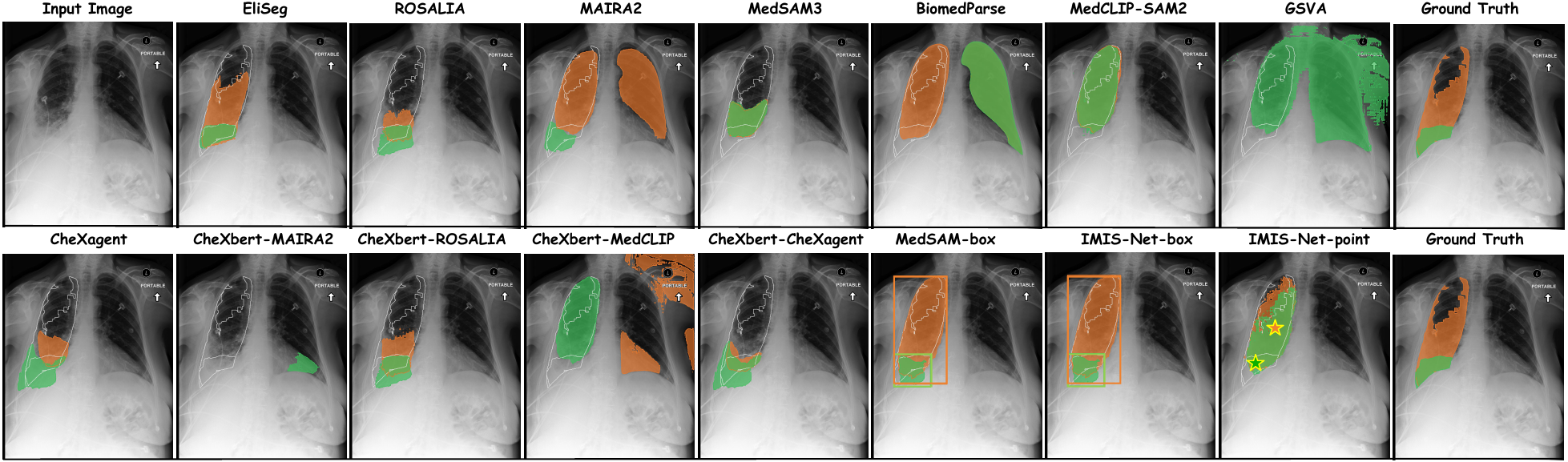}
\caption{Qualitative comparison on a representative multi-finding case.
EliSeg receives the unfiltered report once and jointly predicts masks for
all eligible findings. Each baseline is invoked separately for each target
using its native interface, and the resulting masks are overlaid for
visualization. Orange and green denote the two finding-specific masks, while
white contours indicate the corresponding reference boundaries. Boxes and
points denote the tight spatial prompts used by native spatial
interfaces.}
  \label{fig:qualitative}
\end{figure*}

\paragraph{Overall performance.}

EliSeg achieves the strongest Report-Inferred performance
(Table~\ref{tab:main}). The improvement obtained by
CheXbert$\rightarrow$ROSALIA confirms that target construction is a major
source of error, but also exposes the limitation of a decoupled cascade:
omitted findings cannot be recovered after target extraction, while spurious
findings are propagated directly to mask generation. Figure~\ref{fig:qualitative}
illustrates this distinction on a multi-finding case. EliSeg constructs and
segments both eligible findings in a single inference pass, whereas
target-conditioned baselines require separate runs for each supplied target,
with their masks overlaid for comparison. EliSeg instead verifies the Actor
proposal within the segmentation loop and revises only inconsistent target
structures, preserving the interaction between textual planning and
image-grounded execution.

EliSeg changes only marginally when the gold inventory is supplied, indicating
that most eligibility and cardinality uncertainty has already been resolved
from the report. Its lead in overlap metrics remains under target-specified
inference, showing that the gain extends beyond target verification to the
spatial realization of the selected findings. Under the native-prompt setting, box-prompted methods achieve lower HD95 because the spatial prompt restricts the search region and reduces large boundary deviations. Their lower IoU and Dice, however,
indicate that this advantage does not necessarily translate into more accurate recovery of the complete abnormality extent, whereas EliSeg achieves stronger overall overlap without explicit spatial prompts.

\paragraph{Per-finding performance.}

EliSeg achieves the highest IoU and Dice for every finding in
Table~\ref{tab:per_finding}, showing that the overall improvement is
consistent across categories. Compared with
CheXbert$\rightarrow$ROSALIA, the gain is smallest for cardiomegaly, where
the cascade already performs strongly, and largest for pneumonia and
consolidation. EliSeg also achieves the best or tied NSD for six of the seven
findings and the lowest HD95 for all except pleural effusion. These results
show that its advantage extends beyond target selection to consistent
regional and boundary delineation across findings.

\begin{table}[t]
  \centering
  \begin{threeparttable}
    \small
    \setlength{\tabcolsep}{1pt}
    \renewcommand{\arraystretch}{1.05}

    \begin{tabular}{@{}
      >{\raggedright\arraybackslash}p{0.28\columnwidth}
      *{6}{>{\centering\arraybackslash}
      p{\dimexpr0.12\columnwidth-2\tabcolsep\relax}}
      @{}}
      \toprule
      & \multicolumn{2}{c}{None}
      & \multicolumn{2}{c}{CheXbert}
      & \multicolumn{2}{c}{Qwen-naive} \\
      \cmidrule(lr){2-3}
      \cmidrule(lr){4-5}
      \cmidrule(lr){6-7}

      Backend
    & \mbox{FSR\(\downarrow\)}
    & \mbox{\(\mathrm{IoU}_{+}\uparrow\)}
    & \mbox{FSR\(\downarrow\)}
    & \mbox{\(\mathrm{IoU}_{+}\uparrow\)}
    & \mbox{FSR\(\downarrow\)}
    & \mbox{\(\mathrm{IoU}_{+}\uparrow\)} \\
      \midrule

      GSVA
        & 3.5   & 1.4
        & 1.0   & 1.4
        & 1.7   & 1.4 \\

      MedCLIP-SAMv2
        & 100.0 & 12.2
        & 31.5  & 12.2
        & 58.7  & 12.2 \\

      BiomedParse
        & 100.0 & 15.3
        & 31.5  & 15.3
        & 58.7  & 15.3 \\

      MedSAM3
        & 100.0 & 18.9
        & 31.5  & 18.9
        & 58.7  & 18.9 \\

      CheXagent
        & 98.7  & 28.0
        & 31.5  & 28.0
        & 58.7  & 28.0 \\

      MAIRA-2
        & 74.8  & 31.1
        & 27.3  & 31.1
        & 47.5  & 31.1 \\

      ROSALIA
        & 58.0  & 56.3
        & 24.7  & 56.3
        & 41.3  & 56.3 \\

      \midrule
      \textbf{EliSeg}
        & \multicolumn{6}{c}{
          \(\mathrm{FSR}=14.2\qquad
          \mathrm{IoU}_{+}=60.8\)
        } \\

      \bottomrule
    \end{tabular}

    \caption{Rejection performance on 600 ineligible report mentions.
      FSR is the percentage of ineligible mentions producing a nonempty
      mask, while \(\mathrm{IoU}_{+}\) is IoU on eligible targets.
      External front ends are paired with fixed segmentation backends,
      whereas EliSeg is evaluated as an integrated system. Values are
      reported in percentage points.}

    \label{tab:rejection}
  \end{threeparttable}
\end{table}

\subsection{Rejection Analysis}
\label{sec:rejection}

Report-grounded segmentation must suppress masks for mentions that do not
denote current targets. We therefore report FSR jointly with
\(\mathrm{IoU}_{+}\), since a low FSR can be achieved trivially by
indiscriminately suppressing mask generation.

Table~\ref{tab:rejection} reveals a structural separation between eligibility
control and segmentation capacity. Target-conditioned backends generally
produce a mask once a finding is requested, resulting in high FSR without an
eligibility front end. External report labelers reduce false segmentation but
leave \(\mathrm{IoU}_{+}\) unchanged, while propagating their eligibility
errors directly to the downstream segmenter. EliSeg instead integrates
eligibility verification with mask execution, achieving an FSR of 14.2 while
retaining the highest \(\mathrm{IoU}_{+}\) of 60.8. Although GSVA-based
configurations obtain lower FSR, their \(\mathrm{IoU}_{+}\) of 1.4 indicates
that this apparent selectivity largely reflects limited mask-generation
capacity. Evidence-specific results in Appendix further identify prior mentions as the
main remaining source of rejection error.

\subsection{Ablation Study}
\label{sec:ablation}

\begin{table}[t]
  \centering
  \begin{threeparttable}
    \small
    \begin{tabular}{lccrrrr}
      \toprule
    \multicolumn{1}{c}{\multirow{2}{*}{Config}}
      & \multicolumn{2}{c}{Components}
      & \multicolumn{4}{c}{Metric} \\
      \cmidrule(lr){2-3}\cmidrule(l){4-7}
      & Ver. & Rev.
      & IoU & Dice & NSD & HD95 \\
      \midrule
      w/o Verify & $\times$ & $\surd$ & 49.6 & 66.3 & 16.9 & 426.9 \\
      w/o Revise & $\surd$ & $\times$ & 49.0 & 65.8 & 16.7 & 442.2 \\
      \textbf{EliSeg} & $\surd$ & $\surd$ & \textbf{54.1} & \textbf{72.2} & \textbf{22.9} & \textbf{229.4} \\
      \bottomrule
    \end{tabular}
    \caption{Ablation of the Verifier and Revision using a shared Actor.}
    \label{tab:ablation}
  \end{threeparttable}
\end{table}

As shown in Table~\ref{tab:ablation}, without the Verifier, Revision is
conditioned on the Actor's own target cardinality and therefore tends to
reproduce the original target structure rather than correct it. Without
Revision, the Verifier can detect errors in eligibility and cardinality, but
this information remains at the decision level and cannot alter the number
or content of the predicted masks. The similar performance of the two
partial configurations indicates that neither component is effective in
isolation. Closing the loop improves IoU by 4.5--5.1 points and Dice by
5.9--6.4 points, while substantially reducing HD95. This pronounced boundary
improvement suggests that correcting the target structure helps prevent
severe spatial failures caused by missing, excessive, or mismatched mask
slots. The Verifier thus provides an independent semantic correction signal,
while Revision grounds that signal in new image-based mask predictions.

\begin{table}[t]
  \centering
  \begin{threeparttable}
    \small
    \begin{tabular}{lrrrr}
      \toprule
      Method
      & IoU $\uparrow$
      & Dice $\uparrow$
      & NSD $\uparrow$
      & HD95 $\downarrow$ \\
      \midrule
      IMIS-Net
        & 1.4 & 2.7 & 0.5 & 1245.4 \\
      MedSAM3
        & 10.6 & 19.1 & 4.3 & 434.0 \\
      MedCLIP-SAMv2
        & 11.8 & 21.1 & 3.4 & 437.9 \\
      BiomedParse
        & 12.9 & 22.8 & 6.2 & 368.0 \\
      ROSALIA
        & 29.6 & 45.7 & 6.5 & 391.3 \\
      \textbf{EliSeg (Actor)}
        & \textbf{31.9}
        & \textbf{48.8}
        & \textbf{7.5}
        & \textbf{267.5} \\
      \bottomrule
    \end{tabular}

    \caption{Zero-shot given-target segmentation transfer on CheXlocalize.
Each model receives the gold target through its native text or spatial
interface, without adaptation to CheXlocalize. Because paired reports are
unavailable, EliSeg is evaluated using only its Actor.}

    \label{tab:external}
  \end{threeparttable}
\end{table}

\subsection{Zero-Shot Transfer}
\label{sec:external}

CheXlocalize provides expert masks but no paired radiology reports. We
therefore evaluate zero-shot transfer in a given-target setting, with each
model receiving the reference target through its native interface and no
adaptation to CheXlocalize. For EliSeg, this setting evaluates only the
Actor's segmentation pathway, without the Verifier or Revision.

As shown in Table~\ref{tab:external}, the EliSeg Actor achieves the strongest
performance across all four metrics. Compared with ROSALIA, the improvements
in IoU and Dice are moderate, whereas the reduction in HD95 is substantially
larger. The relatively small NSD gain together with the pronounced HD95
improvement indicates that the main benefit lies in reducing severe boundary
outliers rather than uniformly improving local contour agreement. These results demonstrate effective cross-dataset transfer of the Actor's
image-grounded segmentation capability.

\section{Limitations and Future Work}
\label{sec:limitations}

EliSeg constructs and segments targets directly from unfiltered radiology
reports, but its current scope remains limited. The study covers seven chest
radiographic findings and uses one category-level mask per finding, without
distinguishing bilateral or multifocal instances. Broader vocabularies and
instance-level prediction would improve clinical coverage. The fixed limit
\(K_{\max}=3\) covers most sentences but may truncate rare cases with more
eligible targets, motivating dynamic slot generation. The consistency gate
compares action and cardinality, so semantic errors that preserve the target
count may remain undetected; slot-level verification could provide finer
correction. External evaluation is limited to the Actor because CheXlocalize
lacks paired reports. Evaluating the complete framework will require
multi-institutional datasets with aligned images, reports, and masks.

\section{Conclusion}
\label{sec:conclusion}

We studied abnormality segmentation from radiology reports, where valid
targets must be identified and delineated without prespecified target cues.
We introduced EliSeg, which combines an Actor with independent textual
verification and conditional re-execution to correct target eligibility and
cardinality before final mask decoding. Experiments on MIMIC-CXR-ILS show
that EliSeg improves segmentation across findings while suppressing masks
for ineligible mentions. Ablation studies confirm the complementary roles of
verification and revision, and evaluation on CheXlocalize demonstrates
effective transfer of the Actor to an external dataset. These results show
that EliSeg can construct segmentation targets directly from unfiltered
reports and accurately localize the corresponding abnormalities without
target identities or spatial prompts.

\bibliography{aaai2027}

\setcounter{dbltopnumber}{3}
\renewcommand{\dbltopfraction}{0.95}
\renewcommand{\dblfloatpagefraction}{0.8}
\clearpage
\onecolumn
\vspace*{0.12in}
\begin{center}
  {\LARGE\bfseries Appendix Contents}
\end{center}
\vspace{0.22in}
\hrule
\vspace{0.30in}
{\Large
\noindent\textbf{A\quad Additional Experimental Details}\dotfill 2\par
\vspace{0.35in}
\noindent\hspace*{0.38in}A.1\quad Sentence-Level Annotation\dotfill 2\par
\vspace{0.35in}
\noindent\hspace*{0.38in}A.2\quad Actor Response Grammar\dotfill 2\par
\vspace{0.35in}
\noindent\hspace*{0.38in}A.3\quad Verifier Prompt\dotfill 2\par
\vspace{0.35in}
\noindent\hspace*{0.38in}A.4\quad Target Cardinality and the Choice of $K_{\max}$\dotfill 2\par
\vspace{0.35in}
\noindent\hspace*{0.38in}A.5\quad Training Configuration\dotfill 3\par
\vspace{0.35in}
\noindent\hspace*{0.38in}A.6\quad Model Size\dotfill 3\par
\vspace{0.35in}
\noindent\hspace*{0.38in}A.7\quad Inference Cost\dotfill 5\par
\vspace{0.35in}
\noindent\hspace*{0.38in}A.8\quad Evaluation Protocol\dotfill 6\par
\vspace{0.46in}
\noindent\textbf{B\quad More Results}\dotfill 6\par
\vspace{0.35in}
\noindent\hspace*{0.38in}B.1\quad Evidence-Specific Rejection\dotfill 6\par
\vspace{0.35in}
\noindent\hspace*{0.38in}B.2\quad When Revision Fires, and What It Changes\dotfill 7\par
\vspace{0.35in}
\noindent\hspace*{0.38in}B.3\quad Statistical Considerations\dotfill 8\par
}
\vspace{0.36in}
\hrule
\clearpage
\appendix
\twocolumn[
\begin{center}
    {\LARGE\bfseries Appendix}
\end{center}
\vspace{0.5cm}
]

\section{Additional Experimental Details}

\subsection{Sentence-Level Annotation}
\label{app:dataset}

Each report is split into sentences and each sentence receives one
sentence-level label together with the set of findings it licenses for
segmentation (Table~\ref{tab:data_stats}). A finding becomes a target only when the sentence asserts its current presence \emph{and} a reference mask exists for it. Separating these two conditions yields a fourth state beyond the three actions of Sec.~\ref{sec:problem_formulation}: a sentence that positively asserts a target finding for which no reference mask is available is labelled $\mathrm{UNSUPPORTED}$ and excluded from both training and evaluation. Without this state such sentences would either be supervised toward an empty mask, teaching the model to suppress a genuinely present finding, or be scored as omissions the model had no way to satisfy.

Two distributional properties of the annotation shape the task.
Ineligible mentions are dominated by out-of-scope evidence
($99{,}823$ of $159{,}106$ $\mathrm{REJ}$ sentences in the training split), followed by negated ($34{,}216$), uncertain ($23{,}712$) and prior ($1{,}355$) evidence; prior mentions are thus both the rarest form of ineligible evidence and, as Sec.~\ref{app:evidence_rejection} shows, the hardest to reject. Eligible targets are strongly imbalanced across findings: cardiomegaly accounts
for $36{,}847$ of $88{,}389$ training targets ($41.7\%$), while pneumonia ($2{,}037$) and consolidation ($2{,}921$) together account for under $6\%$. This is a property of how radiologists write rather than of our sampling, and it is why the evaluation universe is balanced across findings instead of following the natural distribution.

\begin{table}[b]
  \centering
  \begin{threeparttable}
    \small
    \setlength{\tabcolsep}{3pt}
    \begin{tabular}{lrrr}
      \toprule
      & \multicolumn{1}{c}{Train}
      & \multicolumn{1}{c}{Val}
      & \multicolumn{1}{c}{Test} \\
      \midrule
      Studies              & 50,160  & 6,078  & 6,505 \\
      Sentences            & 534,703 & 65,313 & 69,434 \\
      \midrule
      \quad $\mathrm{SEG}$          & 80,547  & 9,733  & 10,495 \\
      \quad $\mathrm{REJ}$          & 159,106 & 19,279 & 20,394 \\
      \quad $\mathrm{SILENT}$       & 252,817 & 31,142 & 32,989 \\
      \quad $\mathrm{UNSUPPORTED}$  & 42,233  & 5,159  & 5,556 \\
      \midrule
      Segmentation targets & 88,389  & 10,671 & 11,519 \\
      \bottomrule
    \end{tabular}
    \caption{Sentence-level annotation statistics. Segmentation targets count
      finding-level slots, so a sentence with two eligible findings
      contributes two. The $1{,}008$-target test universe used throughout the
      paper is a balanced subsample of the test split.}
    \label{tab:data_stats}
  \end{threeparttable}
\end{table}

\begin{listing*}[t]
\begin{lstlisting}[numbers=none,breaklines=true]
section: WET READ
"No large effusino or pneumothorax."
  -> REJ  (negated)
     "explicitly states there is no large effusion or pneumothorax"

section: HISTORY
"Interval right upper extremity PICC placement, with tip in mid SVC."
  -> REJ  (out_of_scope)   oos_hits: [picc]

section: HISTORY
"There are small effusions in the lower lobe posteriorly, ? bilateral."
  -> UNSUPPORTED           target: effusion
     (asserted, but no reference mask exists)

section: HISTORY
"The possibility of some underlying collapse and/or consolidation cannot be excluded."
  -> REJ  (uncertain)
     "only hedges the possibility of collapse and consolidation"

section: HISTORY
"Lateral view degraded by respiratory motion."
  -> SILENT

section: HISTORY
"Moderately severe cardiomegaly is unchanged."
  -> SEG   slots: [cardiomegaly]
\end{lstlisting}
\caption{Sentences from a single study, illustrating every label. Note that
\emph{unchanged} is treated as positive rather than prior, since the finding is
still present; and that the misspelling \emph{effusino} is preserved --- reports
are used verbatim, without spelling normalization.}
\label{lst:example}
\end{listing*}

Listing~\ref{lst:example} shows all labels as they occur within one study. It also illustrates why sentence-level eligibility cannot be reduced to keyword matching: the same finding word appears in a negated, an unsupported, and an eligible sentence within a single report, and the deciding evidence is the
surrounding assertion language.

\subsection{Actor Response Grammar}
\label{app:grammar}

The Actor emits a structured response over a compact control vocabulary. The
grammar restricts decoding to the following productions, where $K_{\max}=3$:

\begin{listing*}[t]
\begin{lstlisting}[numbers=none]
<response> ::= <silent> | <reject> | <segment>
<silent>   ::= "None."
<reject>   ::= "It is [REJ]."
<segment>  ::= "It is [SEG1]."
             | "It is [SEG1] and [SEG2]."
             | "It is [SEG1], [SEG2], and [SEG3]."
\end{lstlisting}
\caption{Grammar for the Actor's structured response. A no-segmentation
response denotes $\mathrm{SILENT}$, \texttt{[REJ]} denotes an explicit
rejection, and the $k$-th segmentation token supplies the semantic query for
the $k$-th mask.}
\label{lst:grammar}
\end{listing*}

Segmentation slots follow the canonical finding order
\emph{cardiomegaly, edema, effusion, atelectasis, lung opacity, pneumonia,
consolidation}. Reference targets are serialized in exactly this order during
supervision, so slot index $k$ corresponds to the $k$-th positive finding under
this ordering rather than to an explicitly decoded label. The same ordering is
used by Revision when it constructs a corrected control sequence, which is what
allows slot identities to be recovered without injecting finding names into the
Actor.

\subsection{Verifier Prompt}
\label{app:verifier_prompt}

The Verifier is a frozen Qwen2.5-VL-7B model operating on text only. It
receives a fixed prompt consisting of an instruction block, a schema
declaration, $17$ few-shot examples, and the current input triple
$(\text{section}, h_t, r_t)$. It must return a single JSON object assigning one
state from $\{\mathrm{positive}, \mathrm{negated}, \mathrm{prior},
\mathrm{uncertain}, \mathrm{absent}\}$ to each of the seven target findings,
together with positively asserted out-of-scope mentions and a short rejection
rationale. Decoding is deterministic.

\begin{listing*}[t]
\begin{lstlisting}[numbers=none,breaklines=true]
Classify one sentence from a chest X-ray report. Use section and all prior sentences from that same section as context, but classify the current sentence itself. Section names are context only and must never force a silent result. For every target lesion, return exactly one sentence-level state. A comparison such as "unchanged from prior" is positive when the finding is still present; prior is reserved for resolved or purely historical findings. Add an entry to oos_positive_hits only for a current positive mention of something that is not one of the seven target lesions, never for a negated, prior, or uncertain mention. [...] IMPORTANT: 'history' is provided only to help you resolve references, comparisons, and continuations in the current sentence -- it is NOT additional text to classify. A finding mentioned as positive in history must NOT be carried forward and reported as positive for the current sentence unless the current sentence itself also expresses that finding. Judge strictly from the words actually present in the current sentence -- never use outside clinical knowledge (for example that cardiomegaly commonly co-occurs with pulmonary edema) to infer a finding the sentence itself does not state. A negation before a list such as "no A or B" applies to every item in that list, not just the first one. When a sentence states a definite finding but only hedges its cause or interpretation (phrases like "may represent", "probably due to"), the hedge governs everything after it: mark the underlying observation positive but mark the hedged diagnoses it is attributed to as uncertain, not positive.
Output only one JSON object and no markdown or other text.

SCHEMA:   { target_lesions, allowed_target_states, oos_categories, required_output_keys }
EXAMPLES: 17 input/output pairs
INPUT:    { section, history, sentence }
\end{lstlisting}
\caption{Instruction block of the Verifier prompt (abridged; the schema and
few-shot blocks are serialized as JSON). The prompt asks for a state for all
seven findings rather than for a single queried finding, which is what makes
the verified inventory and its cardinality directly readable from the
response.}
\label{lst:verifier_prompt}
\end{listing*}

Two properties of this design are worth noting. First, the Verifier is asked
to label \emph{every} target finding rather than to answer a query about one
finding, so the eligible inventory $\mathcal{L}_t^V$ and its cardinality
$K_t^V$ follow directly from the response without a separate extraction step.
Second, the instruction explicitly forbids carrying findings forward from
$h_t$ and forbids inferring findings from clinical co-occurrence priors; both
are failure modes we observed when the preceding context is supplied without
such constraints.

\subsection{Target Cardinality and the Choice of $K_{\max}$}
\label{app:cardinality}

Table~\ref{tab:cardinality} gives the distribution of target cardinality over
eligible sentences. The distribution is steeply concentrated: single-target
sentences account for $91.2\%$ of the training split, and $99.92\%$ of eligible
sentences contain at most three targets. Setting $K_{\max}=3$ therefore covers
all but $62$ of $80{,}547$ training sentences, and the corresponding figures for
validation and test are $10$ of $9{,}733$ and $8$ of $10{,}495$. The three
splits agree to within $0.03$ percentage points, so the limit is not an artifact
of the training sample.

\begin{table}[t]
  \centering
  \begin{threeparttable}
    \small
    \setlength{\tabcolsep}{4pt}
    \begin{tabular}{lrrrrrr}
      \toprule
      Split
      & \multicolumn{1}{c}{$K{=}1$}
      & \multicolumn{1}{c}{$K{=}2$}
      & \multicolumn{1}{c}{$K{=}3$}
      & \multicolumn{1}{c}{$K{=}4$}
      & \multicolumn{1}{c}{$K{=}5$}
      & \multicolumn{1}{c}{$K{\leq}3$} \\
      \midrule
      Train & 73,476 & 6,364 & 645 & 60 & 2 & 99.92\% \\
      Val   & 8,911  & 717   & 95  & 9  & 1 & 99.90\% \\
      Test  & 9,573  & 828   & 86  & 8  & 0 & 99.92\% \\
      \bottomrule
    \end{tabular}
    \caption{Target cardinality per eligible sentence. The final column is the
      fraction of eligible sentences fully representable under $K_{\max}=3$.}
    \label{tab:cardinality}
  \end{threeparttable}
\end{table}

\begin{figure}[t]
  \centering
  \includegraphics[width=\columnwidth]{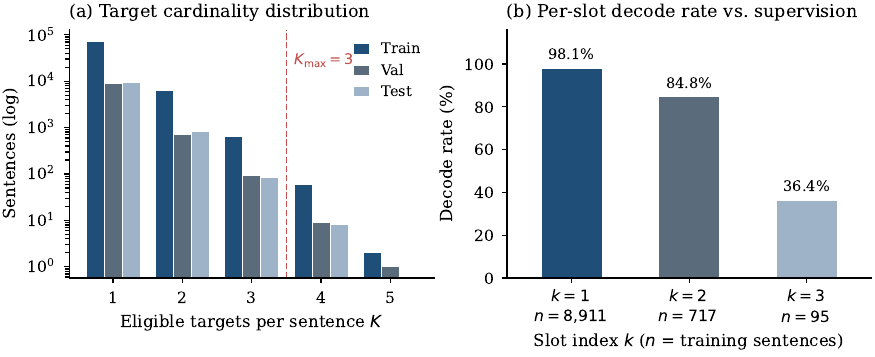}
  \caption{(a) Target cardinality per eligible sentence, on a log scale; the
  three splits agree closely and the mass beyond $K_{\max}=3$ is negligible.
  (b) Per-slot decode rate under autoregressive decoding, against the number of
  training sentences available at that cardinality (right axis, log scale). The
  drop at the third slot follows supervision density, not image difficulty.}
  \label{fig:cardinality}
\end{figure}

The binding constraint is not the grammar limit but supervision density in the
tail. As Fig.~\ref{fig:cardinality}b shows, per-slot decode rates on a
multi-target sample fall from $98.1\%$ at the first slot to $84.8\%$ at the
second and $36.4\%$ at the third, tracking the
number of training sentences available at each cardinality rather than any
property of the images. This is precisely the failure mode Revision addresses:
when the Verifier establishes that $\widetilde{K}_t$ targets are eligible, the
canonical control sequence is constructed directly and every slot is decoded,
so a slot the Actor would have dropped autoregressively still produces a mask.
It also explains why the ablation in Table~\ref{tab:ablation} shows both
partial configurations losing coverage relative to the full loop.

\subsection{Training Configuration}
\label{app:training}

Table~\ref{tab:hparams} lists the full training configuration. The three loss
terms of Eq.~\ref{eq:actor_loss} are weighted
$(\lambda_{\mathrm{CE}}, \lambda_{\mathrm{BCE}}, \lambda_{\mathrm{Dice}})
= (0.5, 5.0, 1.0)$. The comparatively large BCE weight reflects the severe
foreground--background imbalance of finding masks at
$1024\times1024$ resolution; with the Dice term alone, training is dominated by
the largest findings, while BCE supplies a per-pixel gradient that remains
informative for small effusions and focal opacities.

\begin{table}[t]
  \centering
  \begin{threeparttable}
    \small
    \setlength{\tabcolsep}{4pt}
    \begin{tabular}{ll}
      \toprule
      Setting & Value \\
      \midrule
      Initialization        & ROSALIA-7B \\
      Trainable modules     & mask decoder, text projection \\
      Backbone adaptation   & LoRA, rank $8$ \\
      Optimizer             & AdamW \\
      Learning rate         & $5\times10^{-5}$ \\
      Batch size            & $2$ \\
      Gradient accumulation & $4$ steps \\
      Epochs                & $1$ \\
      Image / mask resolution & $1024\times1024$ \\
      $\lambda_{\mathrm{CE}}$ / $\lambda_{\mathrm{BCE}}$ / $\lambda_{\mathrm{Dice}}$
                            & $0.5$ / $5.0$ / $1.0$ \\
      Max slots per sentence & $3$ \\
      Report context        & full preceding same-section text \\
      Hardware              & $1\times$ NVIDIA RTX A6000 \\
      \bottomrule
    \end{tabular}
    \caption{Actor training configuration.}
    \label{tab:hparams}
  \end{threeparttable}
\end{table}

The Verifier is used frozen and is never trained, so it contributes no
gradient and adds no learned parameters to the system. Revision likewise
introduces no parameters: it re-executes the same Actor in inference mode
through a teacher-forced forward pass.

Beyond the mask decoder and text projection, the token embedding and output
head are also updated. This is a consequence of the control vocabulary rather
than a design choice: introducing \texttt{[SEG1]}--\texttt{[SEG3]} and
\texttt{[REJ]} extends the tokenizer, and the embeddings of the new tokens
cannot be learned if the embedding matrix is held fixed. LoRA is applied to
every linear layer of the language backbone, excluding the vision encoder,
multimodal projector, and text projection.

\subsection{Model Size}
\label{app:model_size}

Table~\ref{tab:params} reports parameter counts obtained by traversing each
checkpoint's tensor metadata directly. We report counts rather than checkpoint
file sizes because the released checkpoints mix precisions: ROSALIA-7B and
LISA+ store roughly $0.66$--$0.68$\,B parameters in float32 and the remainder in
bfloat16, so file size ranks them in the opposite order from parameter count.

\begin{table}[t]
  \centering
  \begin{threeparttable}
    \small
    \setlength{\tabcolsep}{4pt}
    \begin{tabular}{lr}
      \toprule
      Component & Params (B) \\
      \midrule
      \multicolumn{2}{l}{\emph{EliSeg Actor, by module}} \\
      \quad Language backbone            & 6.476 \\
      \quad Mask decoder (SAM ViT-H)     & 0.641 \\
      \quad Vision encoder (CLIP ViT-L)  & 0.303 \\
      \quad Token embedding + output head & 0.262 \\
      \quad Text projection              & 0.018 \\
      \quad Multimodal projector         & 0.004 \\
      \quad\textbf{Total}                & \textbf{7.705} \\
      \addlinespace[2pt]
      \quad Trainable                       & 0.921 \\
      \midrule
      \multicolumn{2}{l}{\emph{Reference models}} \\
      \quad ROSALIA-7B                      & 7.402 \\
      \quad GSVA (LISA+ 7B)                 & 7.444 \\
      \quad Verifier (Qwen2.5-VL-7B, frozen) & 8.292 \\
      \quad MedSAM3 (SAM3)               & 0.860 \\
      \quad MedSAM (ViT-B)               & 0.094 \\
      \bottomrule
    \end{tabular}
    \caption{Parameter counts. The EliSeg Actor adds no parameters over its
      ROSALIA-7B initialization; the two agree module for module.}
    \label{tab:params}
  \end{threeparttable}
\end{table}

The comparison with ROSALIA-7B is exact rather than approximate: the two
checkpoints agree module for module --- identical language backbone
($6.476$\,B), mask decoder ($0.641$\,B), embedding and head ($0.262$\,B), text
projection ($0.018$\,B), and projector ($0.004$\,B) --- and the only difference
is that our export bundles the CLIP vision encoder while ROSALIA loads it from
its original release at run time. Under a common accounting both total
$7.705$\,B. EliSeg therefore introduces \emph{no additional parameters} over the
segmenter it is initialized from, and none of the reported improvement can be
attributed to added capacity. The Verifier does add an $8.292$\,B frozen model
at inference time, which we account for separately in
Sec.~\ref{app:inference_cost}, since it is a fixed off-the-shelf component that
is neither trained nor modified.

We note that the mask decoder is a full SAM ViT-H stack; its $0.641$\,B
matches the released SAM ViT-H checkpoint exactly, confirming that EliSeg,
ROSALIA, and GSVA all share the same segmentation backbone and differ only in
how the target is specified to it.

\subsection{Inference Cost}
\label{app:inference_cost}

Table~\ref{tab:latency} reports measured per-stage cost on a single RTX A6000
over $60$ eligible sentences drawn from $44$ studies, after three warmup
iterations, with CUDA synchronization around every timed region. Model loading
is excluded: the checkpoints reside on a single spinning disk, where loading is
dominated by I/O and took $109$--$341$\,s across runs, which characterizes the
storage rather than the model.

\begin{table}[t]
  \centering
  \begin{threeparttable}
    \small
    \setlength{\tabcolsep}{4pt}
    \begin{tabular}{lrr}
      \toprule
      Stage & Latency (ms) & Peak (GiB) \\
      \midrule
      Image encoding, per study      & $593 \pm 107$ & \multirow{4}{*}{16.3} \\
      \quad amortized per sentence   & $435$         & \\
      Actor decoding, per sentence   & $519 \pm 87$  & \\
      Revision forward, per sentence & $548 \pm 68$  & \\
      \midrule
      Verifier, per sentence         & $2868 \pm 194$ & 16.2 \\
      \quad batched ($8$ sentences)  & $1522$         & 21.1 \\
      \midrule
      Actor pathway total   & $1502$ & \\
      Full system, serial   & $4370$ & \\
      Full system, expected & $3942$ & \\
      \bottomrule
    \end{tabular}
    \caption{Measured inference cost. Verifier and Actor run in separate
      processes and environments, so their peak memory figures are not
      additive on one device unless co-located.}
    \label{tab:latency}
  \end{threeparttable}
\end{table}

\begin{figure}[t]
  \centering
  \includegraphics[width=\columnwidth]{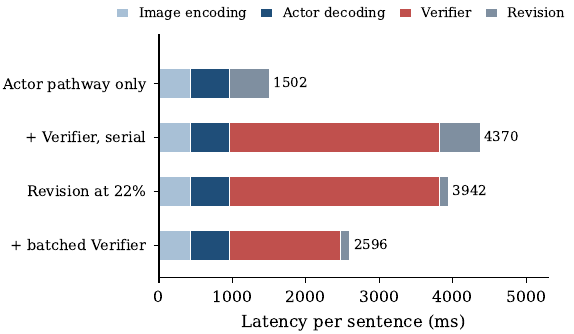}
  \caption{Per-sentence latency decomposition. The image-grounded pathway
  (encoding, Actor, Revision) is a minority of serial end-to-end cost; the
  frozen text Verifier dominates it. Charging Revision only where the gate
  fires, and batching the Verifier's fixed prompt prefix, together bring the
  total below the naive serial figure without changing any prediction.}
  \label{fig:latency}
\end{figure}

\paragraph{Where the cost is.}
As Fig.~\ref{fig:latency} makes clear, the Verifier dominates: at batch size one it accounts for roughly two thirds of
serial end-to-end latency, more than the entire image-grounded pathway. This is
a direct consequence of the prompt design of Sec.~\ref{app:verifier_prompt} ---
the schema and $17$ few-shot examples make the prompt about $13$k characters, so
most of the cost is prefill over a fixed prefix. Two properties follow. First,
that prefix is identical across sentences, so batching amortizes it: eight
sentences per call reduce per-sentence cost to $1522$\,ms, $53\%$ of the
unbatched figure, at the price of $5$\,GiB additional memory. Second, the
Verifier reads only $(h_t, r_t)$ and never observes the image or the Actor
output, so it is independent of the Actor by construction and could be executed
concurrently with it; our implementation runs the two sequentially in separate
environments, and Table~\ref{tab:latency} reports that sequential cost rather
than a projected concurrent one.

\paragraph{Cost of the loop itself.}
The mechanism EliSeg adds over a single forward pass is the Verifier plus a
conditional second Actor execution. The second execution is cheap relative to
the first --- $548$\,ms against $519$\,ms for autoregressive decoding, since it
is a single teacher-forced forward rather than token-by-token generation --- and
it is charged on only $22\%$ of sentences, giving an expected $121$\,ms per
sentence. Image encoding is performed once per study and reused across that
study's sentences, which is what makes the amortized figure ($435$\,ms) lower
than the per-study cost ($593$\,ms); on this sample there were $1.4$ eligible
sentences per study.

\begin{figure}[t]
  \centering
  \includegraphics[width=\columnwidth]{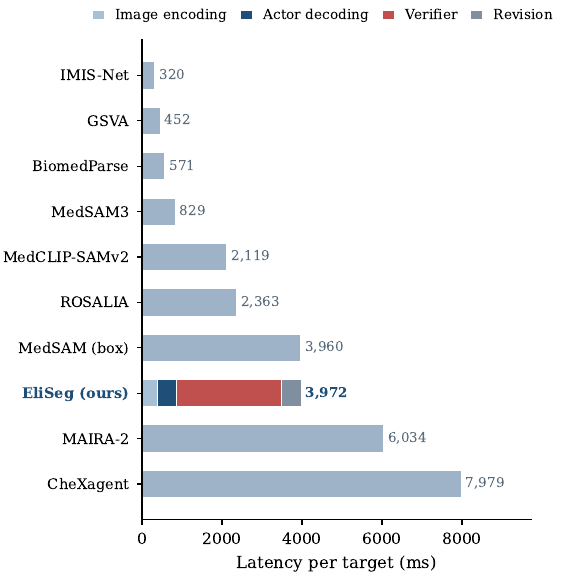}
  \caption{Per-target inference cost. EliSeg is decomposed into its four
  stages; all other methods are single calls. EliSeg sits mid-pack: the
  Verifier accounts for most of its cost, while its image-grounded pathway
  alone is faster than the segmenter it is initialized from.}
  \label{fig:latency_models}
\end{figure}

\paragraph{Comparison to target-conditioned inference.}
Table~\ref{tab:latency_methods} and Fig.~\ref{fig:latency_models} place EliSeg
against every method in Table~\ref{tab:main}, measured under identical
conditions. Because target-conditioned methods are invoked once per target
while EliSeg processes a sentence once regardless of how many findings it
contains, all figures are normalized per target using the $1.1$ targets per
eligible sentence observed on this sample.

\begin{table}[t]
  \centering
  \begin{threeparttable}
    \small
    \setlength{\tabcolsep}{4pt}
    \begin{tabular}{lrrr}
      \toprule
      Method & ms / target & Peak (GiB) & $n$ \\
      \midrule
      IMIS-Net           & 320   & 3.0  & 33 \\
      GSVA               & 452   & ---  & 60 \\
      BiomedParse        & 571   & 2.6  & 60 \\
      MedSAM3            & 829   & 5.3  & 60 \\
      MedCLIP-SAMv2      & 2,119 & 7.1  & 60 \\
      ROSALIA            & 2,363 & 16.0 & 60 \\
      MedSAM (box)       & 3,960 & 2.7\textsuperscript{b} & 60 \\
      \textbf{EliSeg}    & \textbf{3,972} & 16.3 & 60 \\
      \quad Actor pathway only & 1,365 & 16.3 & 60 \\
      MAIRA-2            & 6,034 & 15.5 & 59 \\
      CheXagent          & 7,979 & 14.4 & 60 \\
      \bottomrule
    \end{tabular}
    \caption{End-to-end inference cost per target under identical conditions:
      one RTX A6000, images staged in RAM, warmup discarded, CUDA
      synchronization around each call, model loading excluded. Each figure
      includes that method's own preprocessing, so it measures a deployed call
      rather than an isolated forward pass.}
    \label{tab:latency_methods}
  \end{threeparttable}
\end{table}

Two observations follow. First, EliSeg is not the slowest system here despite
running an extra $8$ B model: it is faster than MAIRA-2 and CheXagent, both of
which generate text autoregressively before segmenting. Second, and more
informative, the EliSeg Actor pathway costs $1{,}365$ ms per target against
ROSALIA's $2{,}363$ ms, even though the two are parameter-identical module for
module (Sec.~\ref{app:model_size}). The difference is structural rather than
architectural: EliSeg encodes each radiograph once and reuses the embedding
across every sentence of that study, whereas a target-conditioned model carries
no cross-sentence state and re-encodes on every invocation. The gap therefore
widens as the number of targets per study grows. The Verifier is what places
EliSeg above ROSALIA in total, and Sec.~\ref{app:inference_cost} above
quantifies how far batching and the $22\%$ trigger rate reduce it.

We report GSVA's latency for completeness, but it should not be read as an
efficiency result: at IoU $=4.6$ (Table~\ref{tab:main}) it rarely produces a
mask at all, and the same caution applies as in Sec.~\ref{sec:rejection}.

\subsection{Evaluation Protocol}
\label{app:protocol}

All systems are scored by a single shared implementation on identical target
universes: $1{,}008$ finding-level targets for MIMIC-CXR-ILS and $899$ for
CheXlocalize. Masks are compared in a common $1024\times1024$ label space.

\paragraph{Overlap metrics.}
IoU and Dice are \emph{pooled} over targets rather than averaged per target.
Writing $P_i$ and $G_i$ for the predicted and reference masks of target $i$,
\begin{equation}
\mathrm{IoU}
=
\frac{\sum_i |P_i \cap G_i|}{\sum_i |P_i \cup G_i|},
\qquad
\mathrm{Dice}
=
\frac{2\sum_i |P_i \cap G_i|}{\sum_i \left(|P_i| + |G_i|\right)}.
\label{eq:pooled}
\end{equation}
Pooling prevents small targets from dominating the aggregate and makes the
metric additive over the universe, so subset results recombine exactly into the
total.

\paragraph{Treatment of missing predictions.}
A target that receives no mask --- because the model rejected it, omitted the
slot, or failed at inference --- is retained as an empty prediction rather than
excluded. Its union term $|P_i \cup G_i| = |G_i|$ therefore remains in the
denominator of Eq.~\ref{eq:pooled}. This is essential for an oracle-free
protocol: a system could otherwise improve its score by declining to predict.
We report coverage, the fraction of targets receiving a nonempty mask, whenever
boundary metrics are discussed.

\paragraph{Boundary metrics.}
NSD uses a tolerance of $5$ pixels. HD95 is reported in pixels; an empty
prediction receives a fixed penalty of $1448.2$ pixels, the diagonal of the
$1024\times1024$ frame, which is the largest attainable distance in that frame.
Boundary metrics cannot be pooled, so both are computed per target and
averaged. Consequently HD95 must be read together with coverage: a
low-coverage system's HD95 is inflated by this penalty rather than by
inaccurate boundaries, which is why methods such as GSVA exhibit HD95 values
above $1300$ in Table~\ref{tab:main} at coverage of $13.6\%$ or lower.

\paragraph{Rejection metrics.}
FSR is evaluated at the level of the individual ineligible finding: a mention
counts as falsely segmented only when the system produces a nonempty mask
\emph{for the specific finding that was injected as ineligible}. A coarser
sentence-level convention, which counts any nonempty mask anywhere in the
sentence, penalizes systems that correctly segment a co-occurring eligible
finding in the same sentence; we report both in
Sec.~\ref{app:evidence_rejection}.

\section{More Results}

\subsection{Evidence-Specific Rejection}
\label{app:evidence_rejection}

Table~\ref{tab:rejection} reports aggregate FSR over $600$ ineligible mentions.
Table~\ref{tab:fsr_evidence} decomposes it by the type of ineligible evidence,
which identifies where the remaining rejection errors are concentrated.

\begin{table}[t]
  \centering
  \begin{threeparttable}
    \small
    \setlength{\tabcolsep}{4pt}
    \begin{tabular}{llrrr}
      \toprule
      Front end & Backend
      & \multicolumn{1}{c}{Neg.}
      & \multicolumn{1}{c}{Prior}
      & \multicolumn{1}{c}{Unc.} \\
      \midrule
      \multirow{5}{*}{None}
        & GSVA           & 2.0   & 3.0   & 5.5 \\
        & MedCLIP-SAMv2\tnote{a}  & 100.0 & 100.0 & 100.0 \\
        & CheXagent      & 96.0  & 100.0 & 100.0 \\
        & MAIRA-2        & 69.0  & 79.5  & 76.0 \\
        & ROSALIA        & 35.5  & 68.5  & 70.0 \\
      \midrule
      \multirow{5}{*}{CheXbert}
        & GSVA           & 0.0   & 2.0   & 1.0 \\
        & MedCLIP-SAMv2\tnote{a}  & 2.5   & 61.0  & 31.0 \\
        & CheXagent      & 2.5   & 61.0  & 31.0 \\
        & MAIRA-2        & 2.5   & 51.0  & 28.5 \\
        & ROSALIA        & 2.0   & 48.0  & 24.0 \\
      \midrule
      \multirow{5}{*}{Qwen-naive}
        & GSVA           & 0.0   & 2.5   & 2.5 \\
        & MedCLIP-SAMv2\tnote{a}  & 14.0  & 88.5  & 73.5 \\
        & CheXagent      & 14.0  & 88.5  & 73.5 \\
        & MAIRA-2        & 12.5  & 70.5  & 59.5 \\
        & ROSALIA        & 9.0   & 61.0  & 54.0 \\
      \midrule
      \multicolumn{2}{l}{\textbf{EliSeg}}
        & \textbf{2.0} & \textbf{29.0} & \textbf{11.5} \\
      \bottomrule
    \end{tabular}
    \begin{tablenotes}[flushleft]
      \footnotesize
      \item[a] BiomedParse and MedSAM3 are identical to MedCLIP-SAMv2 in this
      table: all three segment whatever finding is requested, so their
      false-segmentation behaviour is determined entirely by the front end.
    \end{tablenotes}
    \caption{False-segmentation rate by evidence type, over the same $600$
      ineligible mentions as Table~\ref{tab:rejection} ($200$ per type).
      Values are percentages; lower is better.}
    \label{tab:fsr_evidence}
  \end{threeparttable}
\end{table}

\begin{figure}[t]
  \centering
  \includegraphics[width=\columnwidth]{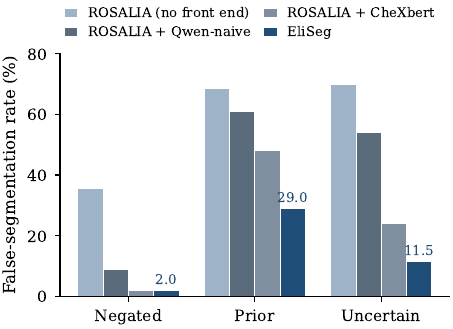}
  \caption{False-segmentation rate by evidence type. Swapping the eligibility
  front end largely resolves negated evidence but leaves prior and uncertain
  evidence poorly handled; EliSeg reduces all three, and the residual is
  concentrated on prior mentions.}
  \label{fig:fsr_evidence}
\end{figure}

Three patterns emerge (Fig.~\ref{fig:fsr_evidence}). First, negated evidence is
easy for any system with an
eligibility front end: explicit negation cues are lexically overt, and every
CheXbert-gated configuration falls below $3\%$. Second, \emph{prior} evidence
is the dominant residual error mode. It remains above $48\%$ for all external
front ends and is the largest of the three for EliSeg as well at $29.0\%$,
which is the observation referenced in Sec.~\ref{sec:rejection}. Distinguishing
a resolved or historical finding from a currently present one requires
temporal reasoning over the sentence rather than cue detection, and the
report often expresses it through comparison language that is
lexically indistinguishable from a present finding. Third, the naive
prompt-based front end is markedly worse than CheXbert on prior and uncertain
evidence while being comparable on negated evidence, indicating that a general
instruction-tuned labeler does not by itself supply temporal assertion
judgement.

EliSeg attains the lowest error on prior and uncertain evidence among all
configurations that retain nontrivial segmentation capacity, and is
essentially tied with the best on negated evidence. The GSVA configurations
are lower still on all three types, but as noted in Sec.~\ref{sec:rejection}
their $\mathrm{IoU}_{+}$ of $1.4$ shows this reflects an inability to produce
masks at all rather than eligibility control.

\begin{figure}[t]
  \centering
  \includegraphics[width=\columnwidth]{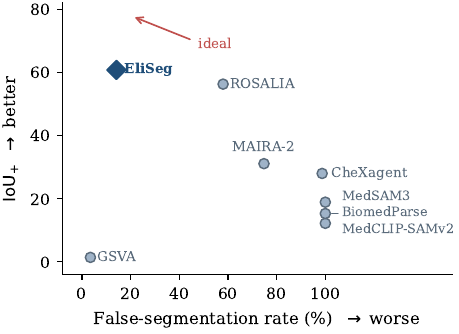}
  \caption{Rejection against segmentation capacity, with no eligibility front
  end. The upper-left corner is desirable: reject ineligible mentions while
  still segmenting eligible ones. GSVA reaches a low false-segmentation rate
  only by producing almost no masks, whereas the remaining
  target-conditioned methods segment whatever is requested. This is why FSR
  must be read jointly with $\mathrm{IoU}_{+}$ rather than on its own.}
  \label{fig:tradeoff}
\end{figure}

Figure~\ref{fig:tradeoff} makes the joint requirement explicit. Reading FSR
alone would rank GSVA first; reading $\mathrm{IoU}_{+}$ alone would hide that
most strong segmenters never decline. Only the two axes together separate
eligibility control from mask-generation capacity, and EliSeg is the sole
configuration that occupies the upper-left region without an external
front end.

\subsection{When Revision Fires, and What It Changes}
\label{app:revision_analysis}

The consistency gate is selective by construction: it re-executes the Actor only
when the Actor and Verifier disagree on action or cardinality. To characterize
what that selectivity buys, we partition a paired sample of $270$ eligible
sentences by the deficit present in the Actor's own output.

\begin{table}[t]
  \centering
  \begin{threeparttable}
    \small
    \setlength{\tabcolsep}{4pt}
    \begin{tabular}{lrr}
      \toprule
      Actor output & \multicolumn{1}{c}{Share} & \multicolumn{1}{c}{$\Delta$ Dice} \\
      \midrule
      Declined to segment       & 10\% & $+0.551$ \\
      Segmented, dropped a slot & 12\% & $+0.216$ \\
      Segmented all slots       & 78\% & $-0.001$ \\
      \bottomrule
    \end{tabular}
    \caption{Effect of Revision by the deficit in the Actor's proposal, on a
      paired sample of $270$ eligible sentences. Rows are disjoint and
      exhaustive. $\Delta$ is measured within each group, so it isolates the
      correction from the composition of the sample.}
    \label{tab:revision_buckets}
  \end{threeparttable}
\end{table}

\begin{figure}[t]
  \centering
  \includegraphics[width=\columnwidth]{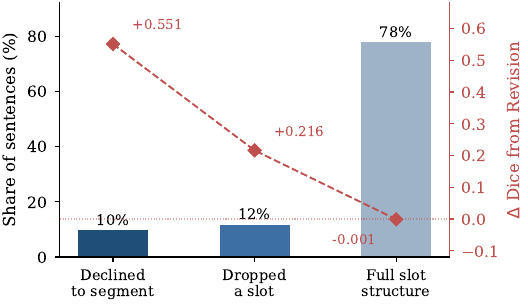}
  \caption{Revision selectively corrects the $22\%$ of sentences with missing target structure while leaving the $78\%$ full-slot group effectively unchanged.}
  \label{fig:revision}
\end{figure}

Two properties of Table~\ref{tab:revision_buckets} and
Fig.~\ref{fig:revision} matter for the claim made in
Sec.~\ref{sec:revision}. First, the gate fires on $22\%$ of sentences, and the
larger share of the total improvement comes from the $10\%$ where the Actor
declined to segment at all --- a deficit no amount of mask refinement could
repair, because there is no mask to refine. Second, on the $78\%$ where the
Actor already produced the full slot structure, Revision is a measured no-op
($-0.001$). This is the property we want from a correction mechanism: it is
evidence that the improvement comes from repairing target structure rather than
from systematically redrawing masks that were already correct. A modification
that also moved the third row would be changing predictions it has no basis to
change.

The gain also concentrates where the Actor's autoregressive decoding is weakest.
Grouping the same sample by reference slot count, Dice improves by $+0.085$ at
one slot, $+0.044$ at two, and $+0.140$ at three or more --- matching the
per-slot decode rates of Sec.~\ref{app:cardinality}, where the third slot is
emitted only $36.4\%$ of the time. The deficit and the correction line up.

\paragraph{Cost of the mechanism.}
Revision is not free. On $330$ eligible sentences the Verifier's proposed
inventory is an exact set match to the reference inventory in $74.2\%$ of cases,
overlaps partially in $23.3\%$, and is disjoint in $2.4\%$; it over-plans on
$5.6\%$ of sentences. Since Revision executes the verified cardinality, an
over-planned inventory materializes a mask that a deployed system would surface
to a reader. The most frequently over-planned finding is lung opacity, which is
also the finding most affected by reference-annotation overlap in
MIMIC-CXR-ILS: a measurable fraction of masks are pixel-identical under
different finding labels among atelectasis, consolidation, opacity, and
pneumonia. Some of these disagreements therefore reflect label arbitrariness in
the reference rather than Verifier error, but we report the raw rate rather than
adjudicate them.

\subsection{Statistical Considerations}
\label{app:stats}

We report deterministic effect sizes rather than null-hypothesis tests. Each
system is trained once and evaluated once on a frozen test universe, so
between-run variance cannot be estimated from the available runs. Treating the
$1{,}008$ targets as independent samples would constitute pseudo-replication:
targets are correlated within a study, findings co-occur systematically
(cardiomegaly with edema, atelectasis with effusion), and several targets in
the universe derive from the same reference mask. A significance test over that
population would therefore report a confidence level that the experimental
design does not support.

Accordingly, all comparisons in this paper are absolute differences in
percentage points on a shared, fixed target universe with a shared metric
implementation. Because the overlap metrics are pooled (Eq.~\ref{eq:pooled}),
these differences are exactly reproducible and decompose additively over
subsets of the universe --- the per-finding rows of
Table~\ref{tab:per_finding} recombine into its \emph{All} row by construction,
and that row reproduces the Report-Inferred column of Table~\ref{tab:main}.
No multiple-comparison correction is needed because we perform no hypothesis
tests.

\end{document}